\documentclass{llncs}
\usepackage[T1]{fontenc}
\usepackage{booktabs}
\usepackage{algorithm}
\usepackage{algpseudocode}
\usepackage{amsmath}
\usepackage{amssymb}
\usepackage[table]{xcolor}
\usepackage{graphicx}
\usepackage{xcolor}
\usepackage{wrapfig}
\usepackage{cite}
\usepackage{multirow}
\usepackage{graphicx}
\usepackage{bbding} 

\newcommand{\comment}[2]{}
\begin{document}
%
\title{GeoSDF-Diff: Geometry-Aware Diffusion Planning for Collision-Free Trajectories}

\titlerunning{GeoSDF-Diff: Geometry-Aware Diffusion Planning}
%
\author{Yueming Li\inst{1}\orcidID{0009-0009-0248-5148
} \and
Zhanjiang Yang\inst{1,2}\orcidID{0009-0009-5349-6111} \and \\
Yang Shen\inst{3}\orcidID{0000-0002-7917-8201
} \and
Meng Li\inst{1}\orcidID{0000-0003-3284-0832
} \and \\
Lijun Sun\inst{1}\orcidID{0000-0002-9830-1527
}\Envelope
}
\authorrunning{Y. Li et al.}
\institute{
Shenzhen Technology University, China \\
\email{202301510025@stumail.sztu.edu.cn, limeng2@sztu.edu.cn, sunlijun@sztu.edu.cn}
\and
Shenzhen University, China \\
\email{2410263030@mails.szu.edu.cn}
\and
University of Technology Sydney, Australia \\
\email{yang.shen-9@student.uts.edu.au}
}

\maketitle              

\begingroup
\renewcommand{\thefootnote}{}
\footnotetext{\Envelope \, Corresponding author.}
\endgroup

\begin{abstract}
Diffusion models have recently attracted increasing attention in motion planning due to their ability to model high-dimensional and multi-modal trajectory distributions. However, existing diffusion-based planners still struggle to reliably satisfy geometric feasibility in unseen cluttered environments with complex obstacle layouts. To address this challenge, we propose \textbf{GeoSDF-Diff}, a geometry-aware diffusion planner that incorporates multi-scale SDF conditioning into trajectory denoising. The key idea is to provide the denoiser with both route-level geometric context and fine local clearance evidence, enabling generated trajectories to better adapt to unseen complex obstacle layouts. Specifically, a Geometric Encoder extracts multi-scale SDF features and injects them into a DiT-X denoiser through cross-attention. We further introduce a trajectory-level training objective that combines SDF clearance, violation emphasis, smoothness, and path-length regularization, so that local infeasibilities are penalized during training rather than corrected only at inference time. Experiments on held-out cluttered environments show that, without post-hoc trajectory repair or sampling-time guidance, GeoSDF-Diff achieves a 65.2\% average success rate across diverse test scenarios, outperforming strong diffusion baselines by 25.4\% to 28.5\%.
\end{abstract}

\section{Introduction}

Motion planning aims to generate collision-free trajectories from start to goal
while satisfying geometric feasibility constraints. This remains challenging in unseen
cluttered environments, where narrow passages, dense obstacles, and complex
free-space connectivity require both global route selection and fine local
clearance reasoning. Classical sampling-based and optimization-based planners
can search or optimize for feasibility, but often depend on extensive collision
checking, good initialization, or iterative cost evaluation
\cite{kavraki2002probabilistic,lavalle1998rapidly,karaman2011sampling,ratliff2009chomp,schulman2014motion,mukadam2018continuous}.
These limitations become more pronounced in cluttered environments,
where feasible regions may be small and difficult to discover efficiently.

Recent learning-based planners, especially diffusion models, offer an
alternative by learning trajectory distributions from data
\cite{ho2020denoising,chi2025diffusion,carvalho2023motion}. Their ability to
model high-dimensional and multi-modal distributions makes them attractive for
generating diverse trajectory candidates. Existing diffusion-based planners
typically improve feasibility through environment conditioning, post-hoc
trajectory refinement, or sampling-time cost guidance
\cite{huang2025diffusionseeder,seo2025presto,carvalho2025motion,luo2024potential}.
While effective, these strategies often do not directly enforce dense geometric
feasibility in the learned denoising process itself.

This gap is critical for collision-free planning. A generated trajectory may be
globally plausible but still violate local clearance constraints near obstacles.
Compact or global map representations can capture coarse obstacle layout, but
may lose fine distance-to-obstacle information required for local collision
avoidance. Conversely, inference-time repair or gradient guidance can improve
feasibility, but introduces additional computation, guidance-weight tuning, and
repeated cost evaluation. An effective diffusion planner should therefore make
geometric information directly available during denoising and penalize local
infeasibility during training.

To address this challenge, we propose \textbf{GeoSDF-Diff}, a geometry-aware
diffusion planner for collision-free trajectory generation in unseen cluttered
environments. GeoSDF-Diff represents trajectories as compact control-point
sequences and converts them into dense trajectories through arc-length
resampling for feasibility evaluation. To make denoising geometry-aware, we
design a Geometric Encoder that combines global SDF map context, start-goal
information, and multi-scale local SDF tokens. Low-resolution SDF features
provide route-level layout and free-space connectivity cues, while
high-resolution features preserve local distance-to-obstacle information. These
conditions are injected into a DiT-X denoiser~\cite{yan2025maniflow} through
cross-attention, enabling trajectory tokens to model inter-waypoint dependencies
while directly attending to geometric evidence. 
We further train the model with a trajectory-level loss that combines dense SDF
clearance, violation emphasis, smoothness, and path-length regularization. The
dense SDF term covers broad clearance errors, while the violation term focuses
on severe local infeasibilities that may be diluted by average penalties.
Experiments on held-out cluttered environments show that
GeoSDF-Diff substantially improves collision-free success over learning-based and
diffusion-based baselines while maintaining efficient amortized inference.
Our main contributions are summarized as follows:
\begin{itemize}
    \item We propose \textbf{GeoSDF-Diff}, a geometry-aware diffusion planning
    framework that generates collision-free trajectories by denoising compact
    trajectory control points under multi-scale SDF conditioning.
    \item We design a Geometric Encoder and a trajectory-level training
    objective to inject multi-scale SDF evidence into denoising and improve
    feasibility-aware trajectory generation.
    \item We demonstrate that GeoSDF-Diff improves the feasibility and
    generalization of diffusion-based planning in held-out cluttered
    environments, achieving stronger collision-free performance without
    post-hoc trajectory repair or sampling-time SDF gradient guidance.
\end{itemize}

\section{Related Work}
\subsection{Classical Motion Planning and Trajectory Optimization}
Classical motion planning methods can be broadly categorized into sampling-based and optimization-based approaches, with sampling-based planners such as PRM~\cite{kavraki2002probabilistic}, RRT~\cite{lavalle1998rapidly}, RRT-Connect~\cite{kuffner2000rrt}, and RRT*~\cite{karaman2011sampling} exploring the collision-free configuration space through random sampling and collision checking. PRM constructs a roadmap by connecting collision-free samples using local planners, whereas RRT-family methods incrementally expand search trees toward sampled states. These methods are broadly applicable and, under appropriate assumptions, some algorithms provide probabilistic completeness, while optimal variants such as RRT* provide asymptotic optimality. However, their computational cost can grow substantially in high-dimensional or cluttered environments, and they are particularly challenged by narrow passages, where extensive sampling and collision checking may be required.

Optimization-based planners formulate motion planning as trajectory optimization over collision, smoothness, and task-related objectives. Representative methods include CHOMP~\cite{ratliff2009chomp}, STOMP~\cite{kalakrishnan2011stomp}, TrajOpt~\cite{schulman2014motion}, GPMP/GPMP2~\cite{mukadam2018continuous}, and CAVE~\cite{tang2024cave}. Compared with sampling-based methods, optimization-based planners can efficiently produce smooth trajectories when provided with a suitable initialization. However, their performance is often sensitive to the initial trajectory and to the nonconvexity of collision-avoidance constraints. Repeated cost evaluation, gradient computation, and constraint checking can further increase computational cost. In challenging cases where the initial trajectory lies far from the feasible region or belongs to an unfavorable homotopy class, local optimization may converge to an infeasible solution or fail to recover a collision-free trajectory.

These limitations motivate the use of learned generative priors to provide higher-quality and more diverse trajectory candidates. Recent diffusion-based planners follow this direction by learning trajectory distributions from data. 
This paper investigates how geometric information and task-specific constraints can be effectively incorporated into the diffusion process to improve the feasibility of generated trajectories.

\subsection{Geometry- and Constraint-Aware Diffusion Planning}

Diffusion models have recently been explored as learned trajectory priors for robot motion planning. DiffusionSeeder~\cite{huang2025diffusionseeder} generates diverse trajectory seeds that are subsequently refined by motion optimization. PRESTO~\cite{seo2025presto} conditions a diffusion model on a key-configuration representation, where a sparse set of task-related configurations is used to approximate C-space obstacles, and then refines the generated trajectories through trajectory optimization. MPD~\cite{carvalho2023motion} takes a different approach by combining a learned trajectory prior with task-specific cost gradients during the reverse denoising process, thereby biasing sampled trajectories toward lower-cost regions at inference time. Potential-based Diffusion Motion Planning~\cite{luo2024potential} further connects diffusion models with potential-based planning by learning trajectory potentials that can compose different motion constraints.

These methods demonstrate the promise of diffusion priors for motion planning, but geometric feasibility is often improved through downstream optimization, compact obstacle representations, or sampling-time guidance. For example, DiffusionSeeder and PRESTO rely on optimization-based refinement to improve feasibility, while MPD requires repeated task-cost evaluation and guidance-weight selection during sampling. In addition, compact C-space representations can encode useful global or task-level obstacle information, but they may not directly provide generated trajectory points with dense local geometric cues, such as distance-to-obstacle, clearance, or nearby collision boundaries.

Our work follows this line of geometry- and constraint-aware diffusion planning, but focuses on making geometric feasibility more directly available to the diffusion model itself. Specifically, we introduce a multi-scale geometry-aware diffusion framework that incorporates SDF-derived geometric information at different spatial resolutions and provides waypoint-level feasibility supervision during training. In this way, the model can exploit both broader obstacle-layout context and local clearance information when generating trajectories. This distinguishes our method from approaches that rely primarily on post-hoc trajectory refinement, compact C-space abstractions, or hand-designed sampling-time guidance, while preserving the generative flexibility of diffusion-based planning.

\section{Method}
\label{sec:method}

We propose \textbf{GeoSDF-Diff}, a geometry-aware diffusion planner for
collision-free trajectory generation in unseen cluttered environments. It
denoises trajectory control points under multi-scale SDF conditioning and uses
trajectory-level losses to improve geometric feasibility.

\subsection{Problem Formulation}
\label{subsec:problem_formulation}

We formulate motion planning as conditional trajectory generation. Given a start
state $\mathbf{s}$, a goal state $\mathbf{g}$, and an SDF map
$\mathbf{S}\in\mathbb{R}^{H\times W}$, the planner generates a collision-free
trajectory from $\mathbf{s}$ to $\mathbf{g}$. For a continuous position
$\mathbf{q}\in\mathbb{R}^{2}$, $\mathbf{S}(\mathbf{q})$ denotes the signed
distance to the nearest obstacle boundary, obtained by bilinear interpolation
from the SDF grid. Positive, zero, and negative values indicate free space,
obstacle boundary, and obstacle interior, respectively.

We represent the trajectory by $K$ intermediate control points,
$\mathbf{x}_0=[\mathbf{c}_1,\ldots,\mathbf{c}_K]\in\mathbb{R}^{K\times 2}$,
where $\mathbf{x}_0$ denotes the clean control-point sequence before diffusion
noise is added. Together with $\mathbf{s}$ and $\mathbf{g}$, these control
points define a piecewise-linear path. The final dense trajectory is obtained
by resampling this path by arc length:
\begin{equation}
    \tau_0
    =
    \tau(\mathbf{x}_0)
    =
    \mathrm{Resample}
    \left(
        [\mathbf{s},\mathbf{c}_1,\ldots,\mathbf{c}_K,\mathbf{g}]
    \right)
    =
    \{\mathbf{q}_i\}_{i=1}^{N}
    \label{eq:resample}
\end{equation}
where $\mathbf{q}_i$ is the $i$-th resampled point, and $N$ is the length of the
dense trajectory.

In particular, we learn the conditional distribution
$p_{\theta}(\mathbf{x}_0 \mid \mathcal{C})$, where
$\mathcal{C}=(\mathbf{S},\mathbf{s},\mathbf{g})$ is the planning condition and
$\theta$ denotes the model parameters. We use a conditional denoising diffusion
probabilistic model (DDPM)~\cite{ho2020denoising} to model this distribution.
At diffusion step $t\in[0,T]$, the forward process corrupts $\mathbf{x}_0$ into a noisy
sequence $\mathbf{x}_t$:
\begin{equation}
q(\mathbf{x}_t \mid \mathbf{x}_0)
=
\mathcal{N}
\left(
\mathbf{x}_t;
\sqrt{\bar{\alpha}_t}\mathbf{x}_0,
(1-\bar{\alpha}_t)\mathbf{I}
\right),
\quad
\bar{\alpha}_t=\prod_{i=1}^{t}(1-\beta_i).
\label{eq:forward_process}
\end{equation}
Here, $\beta_i$ is the noise variance at step $i$ following a predefined noise
schedule, $\bar{\alpha}_t$ is the cumulative signal-retention coefficient, and
$\mathbf{I}$ is the identity matrix. Equivalently, $\mathbf{x}_t$ can be sampled
as
\begin{equation}
\mathbf{x}_t
=
\sqrt{\bar{\alpha}_t}\mathbf{x}_0
+
\sqrt{1-\bar{\alpha}_t}\boldsymbol{\epsilon},
\quad
\boldsymbol{\epsilon}\sim\mathcal{N}(\mathbf{0},\mathbf{I}).
\label{eq:reparameterization}
\end{equation}
Here, $\boldsymbol{\epsilon}$ is standard Gaussian noise. The reverse process
learns to denoise $\mathbf{x}_t$ under the condition $\mathcal{C}$:
\begin{equation}
p_{\theta}(\mathbf{x}_{t-1}\mid \mathbf{x}_t,\mathcal{C})
=
\mathcal{N}
\left(
\mathbf{x}_{t-1};
\boldsymbol{\mu}_{\theta}(\mathbf{x}_t,t,\mathcal{C}),
\sigma_t^2\mathbf{I}
\right),
\label{eq:reverse_process}
\end{equation}
where $\boldsymbol{\mu}_{\theta}$ is the predicted reverse mean and
$\sigma_t^2$ is the reverse variance at step $t$. 
Starting from Gaussian noise, the reverse process generates a clean
control-point sequence $\hat{\mathbf{x}}_0$, which is resampled into the output
dense trajectory $\hat{\tau}_0=\tau(\hat{\mathbf{x}}_0)$.

\begin{figure}[!t]
    \vspace{-1.0em}
    \centering
    \includegraphics[width=\linewidth]{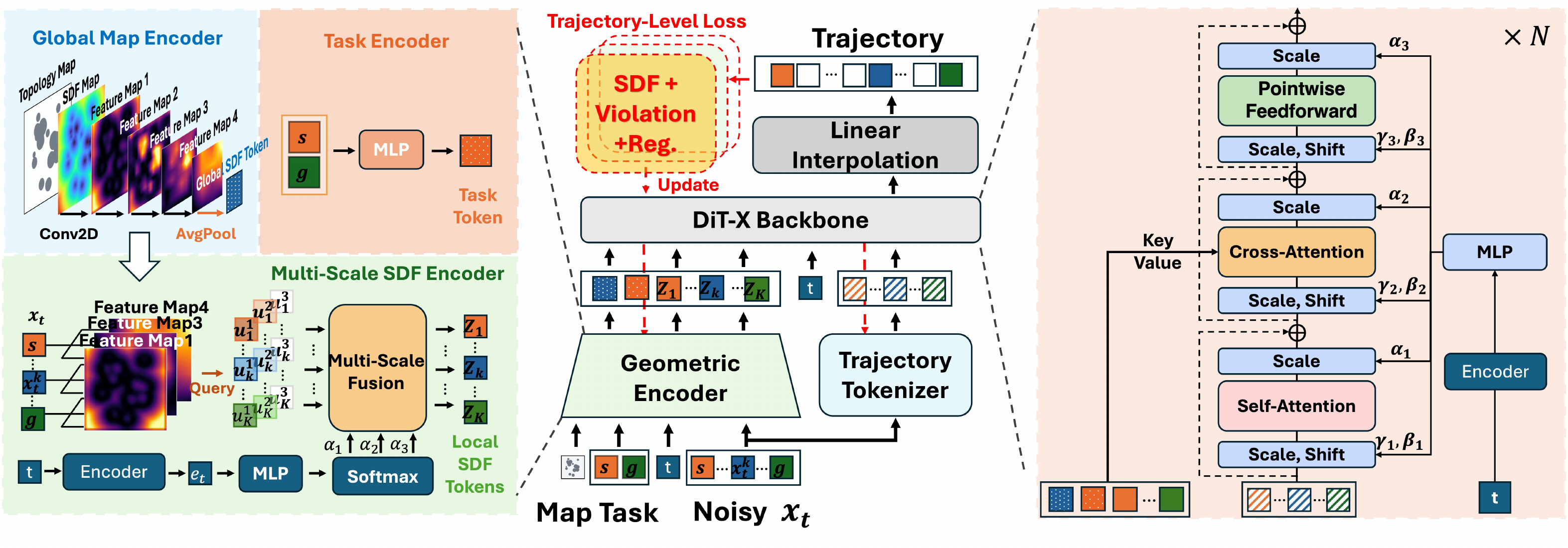}
    \caption{Overview of \textbf{GeoSDF-Diff}, a geometry-aware diffusion planner with multi-scale SDF conditioning. The Geometric Encoder provides global, task, and multi-scale local SDF conditions to a DiT-X denoiser. The predicted control points are resampled into a dense trajectory and supervised by a trajectory-level loss that combines SDF clearance, trajectory-level violation, smoothness, and path-length regularization.}
    \label{fig:framework}
\end{figure}

\subsection{Architecture Overview}
\label{subsec:architecture_overview}

Fig.~\ref{fig:framework} illustrates the overall pipeline of \textbf{GeoSDF-Diff},
a geometry-aware diffusion planner with multi-scale SDF conditioning. Given the
planning condition $\mathcal{C}=(\mathbf{S},\mathbf{s},\mathbf{g})$, the model
starts from Gaussian noise $\mathbf{x}_T$ and progressively denoises it into a
clean control-point sequence $\hat{\mathbf{x}}_0$. The Geometric Encoder extracts
global map context, start-goal information, and multi-scale local SDF tokens,
which condition a DiT-X denoiser through cross-attention. The predicted control
points are then resampled into a dense trajectory
$\hat{\tau}_0=\tau(\hat{\mathbf{x}}_0)$ and supervised by a trajectory-level loss
combining SDF clearance, trajectory-level violation, smoothness, and path-length
regularization.

\subsection{Multi-Scale Geometry-Aware SDF Conditioning}
\label{subsec:multi_scale_sdf}

In cluttered environments, trajectory generation requires both route-level
context and local clearance cues. Low-resolution SDF features capture the
overall obstacle layout and free-space connectivity, while high-resolution
features preserve distance-to-obstacle details. A single global SDF embedding
may fail to retain such multi-scale geometric evidence, making it difficult to
generate collision-free trajectories in cluttered environments. We therefore
design a Geometric Encoder consisting of a Global Map Encoder, a Multi-Scale SDF
Encoder, and a Start-Goal Encoder.

\textbf{Global Map Encoder.}
Given the SDF map $\mathbf{S}$, the Global Map Encoder extracts hierarchical CNN
feature maps at multiple encoder levels. Let
$\mathcal{L}=\{1,\ldots,L\}$ denote the selected set of levels. For each
level $l\in\mathcal{L}$, $E_l(\mathbf{S})$ denotes the corresponding feature
map, and $\pi_l$ projects it to the common hidden dimension $d$, yielding
$\mathbf{F}_l=\pi_l(E_l(\mathbf{S}))$. High-resolution features preserve local
clearance cues, while low-resolution features encode coarser topological
structure.

\textbf{Multi-Scale SDF Encoder.}
At diffusion step $t$, let $\mathbf{x}_t^k\in\mathbb{R}^{2}$ be the $k$-th noisy
control point in $\mathbf{x}_t$. For each level, we query the projected SDF
feature map by bilinear sampling,
$\mathbf{u}_k^l=\mathrm{GridSample}(\mathbf{F}_l,\mathbf{x}_t^k)$. The timestep
embedding $\mathbf{e}_t$ predicts scale weights
$\boldsymbol{\alpha}(t)=\mathrm{softmax}(\mathrm{MLP}(\mathbf{e}_t))$, with
$\sum_{l=1}^{L}\alpha_l(t)=1$. The multi-scale SDF token for the $k$-th control
point is obtained by
\begin{equation}
\mathbf{z}_k(t)
=
\sum_{l=1}^{L}\alpha_l(t)\mathbf{u}_k^l .
\label{eq:multiscale_sdf}
\end{equation}
This timestep-adaptive fusion allows the denoiser to emphasize deeper,
low-resolution SDF features at early noisy steps and gradually shift toward
shallower, high-resolution features for local clearance refinement.

\textbf{Start-Goal Encoder and Condition Sequence.}
We further construct a global SDF token
$\mathbf{z}_{\mathrm{global}}=\mathrm{GAP}(\mathbf{F}_L)$ and a task token
$\mathbf{z}_{\mathrm{task}}=\mathrm{MLP}([\mathbf{s},\mathbf{g}])$, where
$\mathrm{GAP}(\cdot)$ denotes global average pooling. The final geometric
condition sequence is
\begin{equation}
\mathbf{C}_t
=
[
\mathbf{z}_{\mathrm{global}},
\mathbf{z}_{\mathrm{task}},
\mathbf{z}_{1}(t),\ldots,\mathbf{z}_{K}(t)
] .
\label{eq:condition_sequence}
\end{equation}
This sequence provides scene-level geometric context, start-goal information,
and control-point-level SDF evidence to the trajectory denoiser.

\textbf{Trajectory Denoiser.}
We instantiate the diffusion denoiser with DiT-X Transformer
blocks~\cite{yan2025maniflow}. A Trajectory Tokenizer embeds the noisy control
points into trajectory tokens. Each DiT-X block first models dependencies among
control points through self-attention, and then injects the geometric condition
sequence $\mathbf{C}_t$ through cross-attention. The diffusion timestep modulates
the self-attention, cross-attention, and feed-forward branches through adaptive
layer normalization (AdaLN). This design preserves trajectory-sequence reasoning
while making multi-scale SDF information directly available during denoising.

\subsection{Trajectory-level Geometric Constraint Objective}
\label{subsec:sdf_constraint}

The diffusion reconstruction loss encourages accurate denoising, but it does
not explicitly enforce geometric feasibility. We therefore add
reconstruction-level geometric supervision on the predicted clean control-point
sequence $\hat{\mathbf{x}}_0$ and its dense trajectory
$\hat{\tau}_0=\tau(\hat{\mathbf{x}}_0)$. Each loss term targets a different
failure mode: dense SDF supervision improves local clearance, trajectory-level
violation loss prevents severe collisions from being averaged out, and
trajectory regularization encourages smooth and compact paths.

\textbf{Overall Objective.}
The final objective combines the diffusion loss with these geometric terms:
\begin{equation}
    \mathcal{L}_{\mathrm{total}}
    =
    \mathcal{L}_{\mathrm{diff}}
    +
    w_{\mathrm{rec}}
    \left(
        \lambda_{\mathrm{sdf}}\mathcal{L}_{\mathrm{sdf}}
        +
        \lambda_{\mathrm{sm}}\mathcal{L}_{\mathrm{sm}}
        +
        \lambda_{\mathrm{len}}\mathcal{L}_{\mathrm{len}}
        +
        \mathcal{L}_{\mathrm{vio}}
    \right).
    \label{eq:total_loss}
\end{equation}
Here, $\mathcal{L}_{\mathrm{diff}}$ is the standard diffusion reconstruction
loss. The fixed weights $\lambda_{\mathrm{sdf}}$, $\lambda_{\mathrm{sm}}$, and
$\lambda_{\mathrm{len}}$ balance the dense SDF, smoothness, and length losses.
The reconstruction weight $w_{\mathrm{rec}}=\bar{\alpha}_{t}$ assigns stronger
geometric supervision to lower-noise steps, where $\hat{\tau}_0$ is more
reliable for SDF evaluation. Here, $t$ is the diffusion timestep, and
$\bar{\alpha}_{t}=\prod_{i=1}^{t}(1-\beta_i)$ is the cumulative
signal-retention coefficient under the noise schedule $\{\beta_i\}$.

\textbf{Dense SDF Loss.}
This term provides broad local clearance supervision along the whole
trajectory. Let $\hat{\tau}_0=\{\hat{\mathbf{q}}_i\}_{i=1}^{N}$ be the dense
trajectory resampled from $\hat{\mathbf{x}}_0$. We define the clearance
violation at each resampled point as
$r_i=\max(0,d_{\mathrm{safe}}-\mathbf{S}(\hat{\mathbf{q}}_i))$, where
$d_{\mathrm{safe}}$ is the safety clearance margin. The dense SDF loss is
\begin{equation}
    \mathcal{L}_{\mathrm{sdf}}
    =
    \frac{1}{N}
    \sum_{i=1}^{N}
    r_i^2 .
    \label{eq:sdf_loss}
\end{equation}

\textbf{Trajectory-level Violation Loss.}
The dense SDF loss is averaged over many resampled points, so short but severe
collisions may be diluted. To keep such critical infeasibilities visible, we
summarize the dense violations with a trajectory-level residual:
\begin{equation}
    v(\hat{\tau}_0)
    =
    \frac{1}{\gamma_{\mathrm{lse}}}
    \log
    \left(
        \frac{1}{N}
        \sum_{i=1}^{N}
        \exp(\gamma_{\mathrm{lse}} r_i)
    \right).
    \label{eq:trajectory_residual}
\end{equation}
Here, $v(\hat{\tau}_0)$ measures the overall clearance violation of the
trajectory. The parameter $\gamma_{\mathrm{lse}}$ controls the sharpness of this
log-sum-exp soft maximum: a larger value makes the residual closer to the
maximum violation while retaining smooth gradients.

We then penalize this residual with an adaptive trajectory-level violation loss:
\begin{equation}
    \mathcal{L}_{\mathrm{vio}}
    =
    \frac{1}{B}
    \sum_{b=1}^{B}
    \left[
        \lambda_{\mathrm{vio}}
        v(\hat{\tau}_0^{(b)})
        +
        \frac{1}{2}
        v(\hat{\tau}_0^{(b)})^2
    \right].
    \label{eq:violation_loss}
\end{equation}
Here, $B$ is the batch size, $\hat{\tau}_0^{(b)}$ is the dense trajectory of the
$b$-th predicted clean sample, and $\lambda_{\mathrm{vio}}$ is an adaptive
violation weight. A larger $\lambda_{\mathrm{vio}}$ increases the penalty on
persistent trajectory-level violations, while the quadratic term keeps large
residuals strongly penalized.

\textbf{Adaptive Violation Weight.}
The adaptive weight $\lambda_{\mathrm{vio}}$ strengthens the trajectory-level
constraint only when violations persist during training. We compute the detached
batch residual
$\bar{v}=B^{-1}\sum_{b=1}^{B}\mathrm{sg}(v(\hat{\tau}_0^{(b)}))$ and smooth it
as $m_v \leftarrow \gamma_{\mathrm{ema}}m_v+(1-\gamma_{\mathrm{ema}})\bar{v}$,
where $\mathrm{sg}(\cdot)$ denotes stop-gradient and $\gamma_{\mathrm{ema}}$
controls the smoothing rate. The adaptive weight is then updated by
\begin{equation}
\lambda_{\mathrm{vio}}
\leftarrow
\mathrm{clip}
\left(
\lambda_{\mathrm{vio}}
+
\eta_{\lambda}
\max(0,m_v-v_{\mathrm{tol}}),
0,
\lambda_{\max}
\right).
\label{eq:lambda_vio_update}
\end{equation}
Here, $\eta_{\lambda}$ is the update step size, $v_{\mathrm{tol}}$ is the
violation tolerance, and $\lambda_{\max}$ is the upper bound of
$\lambda_{\mathrm{vio}}$. This update avoids reacting to noisy mini-batches
while increasing constraint pressure for persistent infeasibility.

\textbf{Trajectory Regularization Losses.}
Besides collision avoidance, the generated trajectory should also be smooth and
compact. We use the smoothness loss
\begin{equation}
    \mathcal{L}_{\mathrm{sm}}
    =
    \frac{1}{N-2}
    \sum_{i=2}^{N-1}
    \left\|
    \hat{\mathbf{q}}_{i+1}
    -
    2\hat{\mathbf{q}}_i
    +
    \hat{\mathbf{q}}_{i-1}
    \right\|_2^2 ,
    \label{eq:smooth_loss}
\end{equation}
and the length loss
\begin{equation}
    \mathcal{L}_{\mathrm{len}}
    =
    \frac{1}{N-1}
    \sum_{i=1}^{N-1}
    \left\|
    \hat{\mathbf{q}}_{i+1}
    -
    \hat{\mathbf{q}}_i
    \right\|_2 .
    \label{eq:length_loss}
\end{equation}
The former penalizes second-order trajectory variations, while the latter
regularizes path length. Overall, $\mathcal{L}_{\mathrm{sdf}}$ provides dense
clearance supervision, $\mathcal{L}_{\mathrm{vio}}$ emphasizes persistent and
severe trajectory-level violations, and $\mathcal{L}_{\mathrm{sm}}$ and
$\mathcal{L}_{\mathrm{len}}$ improve trajectory regularity.

\section{Experiments}

\subsection{Experimental Setup}

\paragraph{Task.}
We evaluate the proposed method on a 2-D point-robot motion-planning task.
Given a workspace represented by a signed distance field (SDF), the planner is
required to generate a trajectory from a start state to a goal state while
maintaining a prescribed safety clearance from obstacles. We focus on unseen
densely cluttered environments, where narrow passages and limited clearance
make collision-free trajectory generation particularly challenging.

\paragraph{Benchmark Construction.}

\begin{table}[t]
\centering
\caption{Dataset statistics. The Dense-clusters and Square-clusters splits are used only for held-out evaluation. Each generated scene contributes 10 valid
start-goal queries, while the MPD L-R benchmark uses 1000 left-right queries from the MPD scene.}
\label{tab:dataset_statistics}
\begin{tabular}{lccc}
\toprule
Split / Map Type & \# Scenes & Avg. Obstacles & Avg. Occupied Area \\
\midrule
Train / Diagonal-slalom & 256 & 5.02 & 25.80\% \\
Train / Random-clusters & 250 & 5.98 & 21.24\% \\
Train / Two-rooms & 249 & 11.06 & 25.25\% \\
Train / Vertical-wall-gap & 245 & 6.46 & 16.84\% \\
\midrule
Test / Dense-clusters & 200 & 14.97 & 32.61\% \\
Test / Square-clusters & 200 & 13.47 & 40.32\% \\
Test / MPD L-R & 1 & 15.00 & 41.31\% \\
\bottomrule
\end{tabular}
\end{table}

We construct a controlled continuous 2-D benchmark in the workspace
$[-1,1]\times[-1,1]$. Each environment contains multiple circular obstacles and
is represented by a $64\times64$ SDF map, which provides dense geometric
information about obstacle boundaries and free-space clearance of 0.02. Start and goal
states are sampled from opposite sides of the workspace and are accepted only if
both satisfy the safety threshold $d_{\mathrm{safe}}=0.02$.

Expert trajectories are generated by running A$^\ast$ on SDF-valid grid cells,
followed by shortcut smoothing, dense collision checking, and arc-length
resampling into $N=64$ dense points. For diffusion training, we further convert
each dense trajectory into a compact control-point representation by uniformly
selecting $K=16$ intermediate points along its cumulative arc length, while
keeping the start and goal states as separate conditions.

\paragraph{Dataset Splits.}
The benchmark consists of a training split and a held-out dense test split:
\begin{itemize}
    \item \textbf{Training split.} We generate 1,000 scenes from four map families: \emph{vertical-wall-gap}, \emph{diagonal-slalom}, \emph{two-rooms}, and \emph{random-clusters}. Each scene contains 10 valid start-goal pairs, yielding 10,000 training trajectories. On average, the training scenes
    contain 7.12 obstacles and have a 22.33\% occupied-area ratio.
    \item \textbf{Held-out dense test split.}
    We evaluate on \emph{dense-clusters}, \emph{square-clusters}, and the MPD L-R benchmark~\cite{carvalho2023motion}. The generated dense test layouts contain 14.22 obstacles on average and have a 36.47\% occupied-area ratio,     corresponding to about $2.0\times$ more obstacles and a $63.4\%$ higher occupied-area ratio than the training layouts. The MPD L-R scene provides     1,000 additional left-right queries with 15 obstacles and a 41.31\% occupied-area ratio.
\end{itemize}
In total, the held-out evaluation includes 400 generated dense scenes, one MPD scene, and 5,000 test queries. Table~\ref{tab:dataset_statistics} reports the
dataset statistics.

\paragraph{Evaluation Metrics.}
Let $\phi(\mathbf{q})$ denote the SDF value at state $\mathbf{q}$. 
For a trajectory $\tau=\{\mathbf{q}_i\}_{i=1}^{N}$ resampled as in
Eq.~\eqref{eq:resample}, the minimum clearance is defined as
\begin{equation}
d_{\min}(\tau)
=
\min_{1\le i \le N}
\phi(\mathbf{q}_i).
\end{equation}

In particular, we evaluate each generated trajectory using feasibility, quality, and efficiency in terms of the following metrics. 

\begin{itemize}
\item \textbf{Collision rate} is the fraction of trajectories with
$d_{\min}(\tau)<d_{\mathrm{safe}}$.

\item \textbf{Success rate} is the fraction of trajectories satisfying
$d_{\min}(\tau)\ge d_{\mathrm{safe}}$.

\item \textbf{Mean minimum SDF} averages $d_{\min}(\tau)$ over all test
queries, where larger values indicate greater obstacle clearance.

\item \textbf{Mean length} measures the total Euclidean path length,
$\sum_{i=1}^{N-1}\|\mathbf{q}_{i+1}-\mathbf{q}_i\|_2$.

\item \textbf{Mean smoothness} is computed using second-order finite
differences and reflects trajectory curvature.

\item \textbf{Runtime} measures the average planning time per query.

\end{itemize}

\paragraph{Implementation Details.}
All experiments are conducted on a workstation with two NVIDIA RTX PRO 5000 Blackwell GPUs, each with 48 GB memory, dual Intel Xeon Gold 6248R CPUs and 256 GB RAM. The diffusion process uses $T=60$ steps. Unless otherwise specified, we set $\lambda_{\mathrm{sdf}}=0.5$, $\lambda_{\mathrm{sm}}=0.5$, $\lambda_{\mathrm{len}}=0.05$, $\gamma_{\mathrm{lse}}=50$, $\gamma_{\mathrm{ema}}=0.95$, $\eta_{\lambda}=0.02$, $\lambda_{\max}=1000$, and $v_{\mathrm{tol}}=0.005$.

\subsection{Baselines and Variants}


\paragraph{Baselines and Variants.}
We evaluate GeoSDF-Diff, representative baselines, and ablation variants under the same protocol:
\begin{itemize}
    \item \textbf{RRT-Connect$+$GPMP2}~\cite{kuffner2000rrt,mukadam2018continuous}, a classical search-and-optimization pipeline combining sampling-based planning with trajectory optimization.
    \item \textbf{MPD-Bspline}~\cite{carvalho2025motion}, a diffusion-based motion planner that generates B-spline represented trajectory.
    \item \textbf{PRESTO}~\cite{seo2025presto}, a learning-based trajectory generator using key-configuration environment representations.
    \item \textbf{PRESTO$+$cuRobo}~\cite{sundaralingam2023curobo}, a refinement-based variant that applies post-hoc trajectory optimization to PRESTO outputs.
    \item \textbf{GeoSDF-Diff} (ours), the full model with DiT-X denoising, multi-scale SDF conditioning, and the trajectory-level training objective.
    \item \textbf{GeoSDF-Diff w/ DiT} (ours), an ablation that replaces the DiT-X denoising backbone with a standard DiT backbone.
    \item \textbf{GeoSDF-Diff w/o MS}  (ours), an ablation that removes the multi-scale     SDF conditioning and uses a single-scale SDF representation.
    \item \textbf{GeoSDF-Diff w/o $\mathcal{L}_{vio}$} (ours), an ablation that removes the trajectory-level violation loss from the training objective.
\end{itemize}

\subsection{Main Results}

\newcommand{\mvar}[2]{\ensuremath{#1/#2}}
\newcommand{\sdfhead}{SDF/Var}

\begin{table*}[htbp!]
\centering
\caption{Planning comparison: GeoSDF-Diff achieves the best overall performance among diffusion-based methods. The evaluation is conducted. SDF/Var denotes the mean and variance of the minimum SDF, reported in units of $10^{-3}$.}
\label{tab:baseline_sr_sdf_var_clear002}
\setlength{\tabcolsep}{3.2pt}
\renewcommand{\arraystretch}{1.08}
\scriptsize
\resizebox{\textwidth}{!}{
\begin{tabular}{lcccccc|cc}
\toprule
\multirow{2}{*}{\textbf{Method}}
& \multicolumn{2}{c}{\textbf{Dense}}
& \multicolumn{2}{c}{\textbf{Square}}
& \multicolumn{2}{c}{\textbf{MPD L-R}}
& \multicolumn{2}{c}{\textbf{Average}} \\
\cmidrule(lr){2-3}
\cmidrule(lr){4-5}
\cmidrule(lr){6-7}
\cmidrule(lr){8-9}
& SR $\uparrow$ & \sdfhead
& SR $\uparrow$ & \sdfhead
& SR $\uparrow$ & \sdfhead
& SR $\uparrow$ & \sdfhead \\
\midrule

RRT-Connect$+$GPMP2
& \textbf{100.0} & \textbf{\mvar{0.059}{0.35}}
& \textbf{99.9} & \textbf{\mvar{0.049}{0.15}}
& \textbf{98.9} & \textbf{\mvar{0.059}{0.30}}
& \textbf{99.6} & \textbf{\mvar{0.055}{0.27}} \\

MPD-Bspline
& 47.2 & \mvar{0.005}{1.72}
& 21.8 & \mvar{-0.008}{0.89}
& 50.3 & \mvar{0.015}{0.72}
& 39.8 & \mvar{0.004}{1.11} \\

PRESTO
& 38.5 & \mvar{0.002}{1.55}
& 20.3 & \mvar{-0.008}{0.91}
& 51.3 & \mvar{0.011}{1.44}
& 36.7 & \mvar{0.002}{1.30} \\

PRESTO$+$cuRobo
& 37.2 & \mvar{0.003}{1.59}
& 20.5 & \mvar{-0.007}{0.92}
& 50.2 & \mvar{0.011}{1.48}
& 36.0 & \mvar{0.002}{1.33} \\

\midrule
GeoSDF-Diff w/o MS
& 57.5 & \mvar{0.015}{1.31}
& 27.6 & \mvar{0.000}{0.74}
& \underline{72.3} & \mvar{0.026}{0.50}
& 52.5 & \mvar{0.014}{0.85} \\

GeoSDF-Diff w/o $\mathcal{L}_{\mathrm{vio}}$
& \underline{86.2} & \mvar{0.029}{0.54}
& 37.4 & \mvar{0.009}{0.53}
& 58.5 & \mvar{0.020}{0.92}
& 60.7 & \mvar{0.019}{0.67} \\

\rowcolor{gray!12}
\textbf{GeoSDF-Diff (ours)}
& 86.1 & \mvar{0.030}{0.60}
& \underline{38.6} & \mvar{0.009}{0.56}
& 71.1 & \mvar{0.024}{0.66}
& \underline{65.2} & \mvar{0.021}{0.60} \\
\bottomrule
\end{tabular}
}
\end{table*}

\newcommand{\smoothhead}{Smooth. $\downarrow$ }

\begin{table*}[t]
\centering
\caption{Trajectory quality and runtime comparison on Dense, Square, and MPD L-R. GeoSDF-Diff achieves competitive trajectory quality with efficient runtime. We compare trajectory length (L), smoothness (Sm), and runtime (T); lower values are better for all metrics. Smoothness values are scaled by $10^{-4}$, and RRT+GPMP2 denotes RRT-Connect+GPMP2.}
\label{tab:baseline_length_smooth_time_clear002}
\setlength{\tabcolsep}{1.4pt}
\renewcommand{\arraystretch}{1.04}
\scriptsize
\begin{tabular}{@{}l*{9}{r}@{\hspace{4pt}}|@{\hspace{4pt}}*{3}{r}@{}}
\toprule
\multirow{2}{*}{\textbf{Method}}
& \multicolumn{3}{c}{\textbf{Dense}}
& \multicolumn{3}{c}{\textbf{Square}}
& \multicolumn{3}{c}{\textbf{MPD L-R}}
& \multicolumn{3}{c}{\textbf{Average}} \\
\cmidrule(lr){2-4}
\cmidrule(lr){5-7}
\cmidrule(lr){8-10}
\cmidrule(lr){11-13}
& L & Sm & T
& L & Sm & T
& L & Sm & T
& L & Sm & T \\
\midrule
RRT+GPMP2
& 2.665 & 7.220 & 0.230
& 2.614 & 9.540 & 0.235
& 2.584 & 7.560 & 0.285
& 2.621 & 8.110 & 0.250 \\
MPD-Bspline
& \underline{2.087} & 0.449 & 0.870
& \textbf{2.047} & 0.422 & 0.927
& 2.060 & 0.400 & 0.895
& \underline{2.065} & 0.424 & 0.898 \\
PRESTO
& 2.166 & 4.330 & \textbf{0.030}
& 2.123 & 4.650 & \textbf{0.030}
& 2.112 & 2.810 & \textbf{0.030}
& 2.134 & 3.930 & \textbf{0.030} \\
PRESTO+cuRobo
& 2.103 & 6.210 & 0.135
& \underline{2.063} & 6.250 & 0.112
& 2.050 & 6.100 & 0.134
& 2.072 & 6.190 & 0.127 \\
\midrule
GeoSDF. w/o MS
& 2.114 & 0.103 & \underline{0.041}
& 2.084 & \textbf{0.144} & \underline{0.042}
& 2.062 & 0.127 & \underline{0.042}
& 2.087 & 0.125 & \underline{0.042} \\
GeoSDF. w/o $\mathcal{L}_{\mathrm{vio}}$
& \textbf{2.077} & \textbf{0.089} & 0.057
& 2.098 & 0.157 & 0.057
& \textbf{2.002} & \textbf{0.117} & 0.057
& \textbf{2.059} & \textbf{0.121} & 0.057 \\
\rowcolor{gray!12}
\textbf{GeoSDF-Diff}
& 2.101 & \underline{0.092} & 0.057
& 2.101 & \underline{0.155} & 0.057
& \underline{2.034} & \underline{0.121} & 0.057
& 2.078 & \underline{0.123} & 0.057 \\
\bottomrule
\end{tabular}
\end{table*}

\textbf{Geometric feasibility.}
Table~\ref{tab:baseline_sr_sdf_var_clear002} evaluates geometric feasibility on held-out cluttered maps. GeoSDF-Diff achieves the best average feasibility among learned and diffusion-based planners, improving the average success rate from $39.8$\% for MPD-Bspline and $36.7$\% for PRESTO to $65.2$\%. This result supports the main hypothesis of our method: making SDF-derived geometric evidence directly available during denoising helps diffusion planners better handle unseen cluttered environments. In particular, the multi-scale SDF tokens provide both route-level map context and local clearance cues, while the SDF-conditioned denoiser injects this information into trajectory token updates. The higher mean minimum SDF ($0.021$ on average) further indicates that the improvement is not merely due to a higher success rate, but also corresponds to better local clearance along the generated trajectories. 

\textbf{Comparison with search-and-optimization.}
RRT-Connect$+$GPMP2 remains the strongest feasibility baseline, reaching $99.6$\% average success. This is expected because it explicitly searches for a feasible path and then optimizes the trajectory. In contrast, GeoSDF-Diff is an amortized generative planner that samples trajectories from the learned denoising model without post-hoc trajectory repair or sampling-time SDF-gradient guidance. The comparison therefore highlights a different operating point: explicit search-and-optimization remains preferable when maximum feasibility is the only criterion, whereas GeoSDF-Diff substantially improves the feasibility of diffusion-based planning while preserving direct generative inference.

\textbf{Trajectory quality and efficiency.}
Table~\ref{tab:baseline_length_smooth_time_clear002} further shows that the feasibility gains do not come from generating excessively long, irregular, or slow trajectories. GeoSDF-Diff keeps path length close to the learned and diffusion-based baselines, with an average length of $2.078$, while producing substantially smoother trajectories than PRESTO, PRESTO$+$cuRobo, and RRT-Connect $+$ GPMP2. Its runtime remains in the amortized inference regime at $0.057$s on average, much lower than MPD-Bspline and RRT-Connect$+$GPMP2, although PRESTO is faster. Overall, these results show that GeoSDF-Diff improves geometric feasibility while maintaining practical trajectory quality and efficient direct inference.

\begin{figure}
    \centering
    \includegraphics[width=\linewidth]{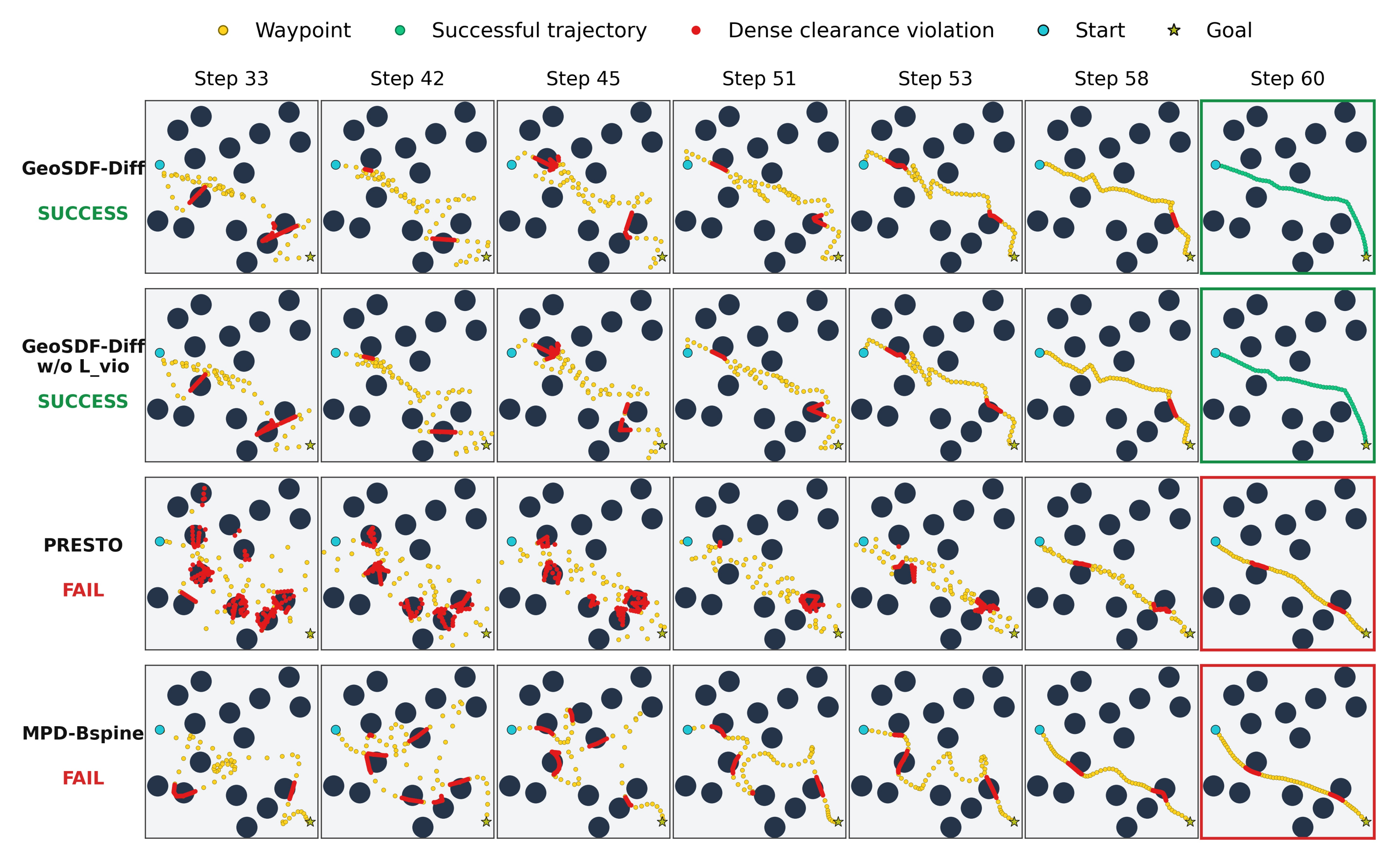}
    \caption{Denoising trajectories on an MPD simple test case. Columns show selected reverse diffusion steps, red points denote dense clearance violations, and green points indicate successful   final trajectory.}
    \vspace{-2em}
    \label{fig:visualization}
\end{figure}

\textbf{Qualitative visualization.}
Fig.~\ref{fig:visualization} provides a qualitative view of the reverse denoising process on an unseen MPD L-R case. GeoSDF-Diff progressively reduces dense clearance violations and produces a collision-free final trajectory, while PRESTO and MPD-Bspline retain local violations in the final output. This visualization is consistent with the quantitative results in Table~\ref{tab:baseline_sr_sdf_var_clear002}, suggesting that SDF-conditioned denoising helps correct local infeasibilities before the trajectory is finalized.

\subsection{Ablation Study}

\begin{table*}[htbp]
\centering
\caption{\textbf{Ablation study}.
We evaluate the effects of DiT-X backbone, multi-scale SDF features (MS), and the trajectory-level violation objective. Success rate (SR) and SDF/Var are reported, where SDF/Var denotes the mean minimum SDF and its variance scaled by $10^{-3}$.}
\label{tab:ablation_sr_sdf_var}
\setlength{\tabcolsep}{3.2pt}
\renewcommand{\arraystretch}{1.08}
\scriptsize
\resizebox{\textwidth}{!}{
\begin{tabular}{lcccccccc}
\toprule
\textbf{Method}
& \multicolumn{2}{c}{\textbf{Dense}}
& \multicolumn{2}{c}{\textbf{Square}}
& \multicolumn{2}{c}{\textbf{MPD L-R}}
& \multicolumn{2}{c}{\textbf{Average}} \\
\cmidrule(lr){2-3}
\cmidrule(lr){4-5}
\cmidrule(lr){6-7}
\cmidrule(lr){8-9}
& SR $\uparrow$ & \sdfhead
& SR $\uparrow$ & \sdfhead
& SR $\uparrow$ & \sdfhead
& SR $\uparrow$ & \sdfhead \\
\midrule
GeoSDF-Diff w/ DiT
& 60.5 & \mvar{0.018}{1.23}
& 30.7 & \mvar{0.004}{0.67}
& 54.9 & \mvar{0.017}{0.77}
& 48.7 & \mvar{0.013}{0.89} \\

GeoSDF-Diff w/o MS
& 57.5 & \mvar{0.015}{1.31}
& 27.6 & \mvar{0.000}{0.74}
& \textbf{72.3} & \textbf{\mvar{0.026}{0.50}}
& 52.5 & \mvar{0.014}{0.85} \\

GeoSDF-Diff w/o $\mathcal{L}_{\mathrm{vio}}$
& \textbf{86.2} & \mvar{0.029}{0.54}
& \underline{37.4} & \mvar{0.009}{0.53}
& 58.5 & \mvar{0.020}{0.92}
& \underline{60.7} & \mvar{0.019}{0.67} \\

\rowcolor{gray!12}
\textbf{GeoSDF-Diff (ours)}
& \underline{86.1} & \textbf{\mvar{0.030}{0.60}}
& \textbf{38.6} & \textbf{\mvar{0.009}{0.56}}
& \underline{71.1} & \mvar{0.024}{0.66}
& \textbf{65.2} & \textbf{\mvar{0.021}{0.60}} \\
\bottomrule
\end{tabular}
}
\end{table*}

\begin{figure}[htbp]
    \vspace{-3.0em}
    \centering
    \includegraphics[width=0.9\linewidth]{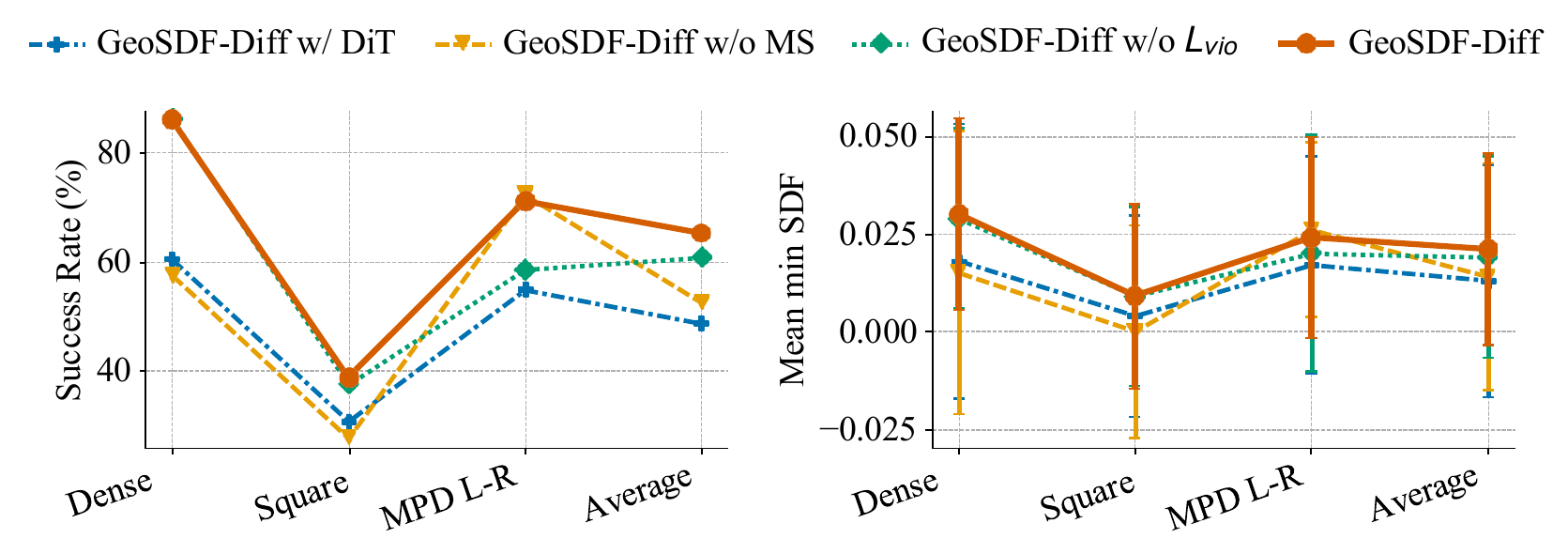}
    \caption{Ablation of the SDF-conditioned DiT-X denoiser, multi-scale SDF
    conditioning, and trajectory-level violation objective.}
    \label{fig:ablation}
\end{figure}

\textbf{Effect of SDF-conditioned denoising.}
Table~\ref{tab:ablation_sr_sdf_var} isolates the contributions of the SDF-conditioned DiT-X denoiser, the multi-scale SDF representation, and the trajectory-level violation objective. Replacing DiT-X with a standard DiT backbone reduces the average success rate from $65.2$\% to $48.7$\% and the mean minimum SDF from $0.021$ to $0.013$. This indicates that a generic diffusion backbone over trajectory control points is insufficient for geometry-sensitive planning. Injecting SDF and start-goal condition tokens through cross-attention provides task-relevant geometric evidence during trajectory denoising.

\textbf{Effect of multi-scale SDF conditioning.}
Removing multi-scale SDF conditioning reduces the average success rate to $52.5$\%, with clear drops on Dense and Square. These results indicate that cluttered layouts require both coarse route-level structure and fine local clearance cues. Although the single-scale variant performs slightly better on MPD L-R, its weaker average performance suggests lower robustness across diverse held-out environments.

\textbf{Effect of trajectory-level violation supervision.}
Removing the adaptive trajectory-level violation objective lowers the average success rate to $60.7$\% and is especially harmful on MPD L-R, where the success rate decreases from $71.1$\% to $58.5$\%. This suggests that dense pointwise SDF supervision alone may under-penalize short but severe local collisions, since such violations can be diluted when averaged over the full trajectory. The trajectory-level violation term keeps these infeasibilities explicit during training.

Overall, the full GeoSDF-Diff model achieves the best average feasibility among the GeoSDF-Diff variants. The ablation trends in Table~\ref{tab:ablation_sr_sdf_var} and Fig.~\ref{fig:ablation} further show that SDF-conditioned denoising, multi-scale geometric representation, and trajectory-level feasibility supervision contribute complementary benefits to geometry-aware diffusion planning.

\section{Conclusion}

We presented \textbf{GeoSDF-Diff}, a geometry-aware diffusion planner for collision-free trajectory generation in unseen cluttered environments. By combining multi-scale SDF conditioning, an SDF-conditioned DiT-X denoiser, and trajectory-level geometric supervision, GeoSDF-Diff enables the diffusion model to exploit both global obstacle-layout context and local clearance cues during denoising, reducing its reliance on post-hoc trajectory repair or sampling-time cost-gradient guidance. Experiments on diverse held-out test scenarios show that GeoSDF-Diff achieves a $65.2\%$ average success rate, outperforming MPD-Bspline and PRESTO by $25.4$\% and $28.5$\%, respectively, while maintaining efficient amortized inference and smooth trajectory generation. Ablation results further confirm the complementary benefits of DiT-X conditioning, multi-scale SDF features, and the trajectory-level violation objective. Although search-and-optimization methods still achieve higher absolute feasibility, GeoSDF-Diff demonstrates the potential of efficient geometry-aware generative motion planning in unseen cluttered 2-D environments, and extending the framework to 3-D environments and higher-dimensional robotic configuration spaces is an important direction for future work.




\begin{credits}

\subsubsection{\ackname}
This work was supported by Natural Science Foundation of Top Talent of SZTU
under Grant No. GDRC202521.

\subsubsection{\discintname}
The authors have no competing interests to declare that are relevant to
the content of this article.

\end{credits}

%
%
%

\bibliographystyle{splncs04}
\bibliography{reference}
%




\end{document}